\pdfoutput=1

\documentclass[11pt]{article}

\usepackage[]{acl}
\usepackage{times}
\usepackage{latexsym}

\usepackage[T1]{fontenc}

\usepackage[utf8]{inputenc}

\usepackage{microtype}

\usepackage{adjustbox}
\usepackage{booktabs}
\usepackage{textcomp}
\usepackage{xcolor}

\newcommand{\draftonly}[1]{#1}

\newcommand{\draftcomment}[1]{\draftonly{#1}}

\newcommand{\zhilin}[1]{\draftcomment{{\color{purple}[#1]$_{zw}$}}}

\usepackage{arydshln}  

\title{Extracting and Inferring Personal Attributes from Dialogue}

\author{
  Zhilin Wang*, Xuhui Zhou, Rik Koncel-Kedziorski, Alex Marin, Fei Xia\\
  University of Washington, *Nvidia\\
  \texttt{\{zhilinw, xuhuizh, kedzior, amarin, fxia\}@uw.edu}\\
}
\date{}

\begin{document}
\maketitle

\begin{abstract}

  Personal attributes represent structured information about a person, such as their hobbies, pets, family, likes and dislikes. We introduce the tasks of extracting and inferring personal attributes from human-human dialogue, and analyze the linguistic demands of these tasks. To meet these challenges, we introduce a simple and extensible model that combines an autoregressive language model utilizing constrained attribute generation with a discriminative reranker. Our model outperforms strong baselines on extracting personal attributes as well as inferring personal attributes that are not contained verbatim in utterances and instead requires commonsense reasoning and lexical inferences, which occur frequently in everyday conversation. Finally, we demonstrate the benefit of incorporating personal attributes in social chit-chat and task-oriented dialogue settings. 
\end{abstract}



\section{Introduction}

\begin{figure}[h]
        \centering
        \includegraphics[width=8cm]{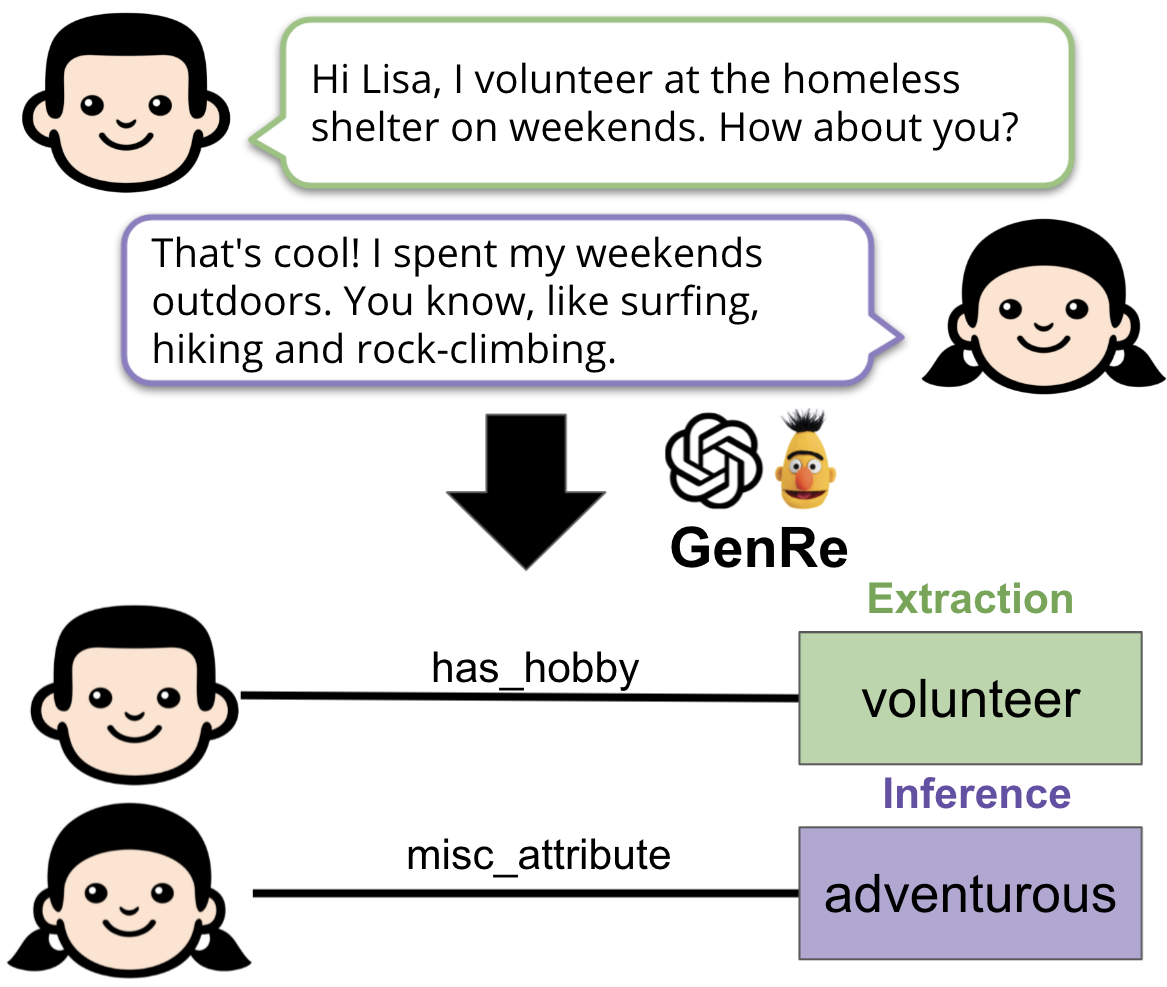}
        \caption{Overview of obtaining personal attribute triple from utterances using our model GenRe. Attribute values are contained within the utterance in the \textsc{Extraction} task, but not the \textsc{Inference} task.}
        \label{fig:front_figure}
\end{figure}

Personal attributes are structured information about a person, such as what they like, what they have, and what their favorite things are. These attributes are commonly revealed either explicitly or implicitly during social dialogue as shown in Figure \ref{fig:front_figure}, allowing people to know more about one another. These personal attributes, represented in the form of knowledge graph triples (\emph{e.g.} I, has\_hobby, volunteer), can represent large numbers of personal attributes in an interpretable manner, facilitating their usage by weakly-coupled downstream dialogue tasks \citep{7078578, ijcai2018-0595, zheng2020personalized, Zheng_Zhang_Huang_Mao_2020, hogan2021knowledge}. 

One such task is to ground open-domain chit-chat dialogue agents to minimize inconsistencies in their language use (\emph{e.g.}, \textbf{I like cabbage} \textrightarrow (next turn) \textrightarrow \textbf{Cabbage is disgusting}) and make them engaging to talk with \citep{ li-etal-2016-persona, zhang-etal-2018-personalizing, mazare-etal-2018-training, ijcai2018-0595, zheng2020personalized, Zheng_Zhang_Huang_Mao_2020, Li_Jiang_Feng_Sprague_Zhou_Hoey_2020,  majumder-etal-2020-like}. 
Thus far, personalization in chit-chat has made use of dense embeddings and natural language sentences. While KG triples have been shown to be capable of grounding Natural Language Generation \citep{moon-etal-2019-opendialkg, koncel-kedziorski-etal-2019-text}, they have yet to be used to personalize chit-chat dialogue agents.

Personal attributes can also help task-oriented dialogue agents to provide personalized recommendations \citep{ mo2017personalizing, joshi2017personalization, Luo_Huang_Zeng_Nie_Sun_2019, lu-etal-2019-goal, pei2021cooperative}. Such personalized recommendations have only been attempted for single-domain tasks with a small set of one-hot features ($<30$). Personalization across a wide range of tasks (recommending food, movies and music by multi-task dialogue agents such as Alexa, Siri and Assistant) however can require orders of magnitude more personal attribute features. This makes KG triples ideal for representing them, given the advantages of this data structure for models to select and utilize pertinent features \citep{7078578, hogan2021knowledge}. 

Based on these advantages, we investigate how personal attributes can be predicted from dialogue. An important bottleneck for this step lies in the poor coverage of relevant personal attributes in existing labeled datasets. Therefore, we introduce two new tasks for identifying personal attributes in Section \ref{sec:pa_task}. As shown in Figure \ref{fig:front_figure}, the \textsc{Extraction} task requires determining which phrase in an utterance indicate a personal attribute, while the \textsc{Inference} task adds further challenge by requiring models to predict personal attributes that are not explicitly stated verbatim in utterances. This is common in conversational settings, where people express personal attributes using a variety of semantically related words or imply them using commonsense reasoning. We analyze how these factors allow personal attributes to be linked to utterances that express them.


To tackle these tasks, we propose a simple yet extensible model, {\bf GenRe}, in Section \ref{sec:genre}. GenRe combines a constrained attribute generation model (that is flexible to accommodate attributes not found verbatim in utterances) with a discriminative reranker (that can contrast between highly similar candidates). Our experiments in Section \ref{sec:experiments} suggest that such design allows our model to outperform strong baseline models on both the \textsc{Extraction} and \textsc{Inference} tasks. Subsequently in Section \ref{sec:overall_analysis}, detailed ablation studies demonstrate the value of our model components while further analysis identifies future areas for improvement.

Finally in Section \ref{sec:benefits_to_personachat}, we show how personal attributes in the form of KG triples can improve the personalization of open-domain social chit-chat agents as well as task-oriented dialogue agents. In the former case, personal attributes can be utilized to improve chat-bot consistency on the PersonaChat task \citep{zhang-etal-2018-personalizing}. In the latter case, we suggest how our personal attributes can support personalization in multi-task, task-oriented dialogue settings.

\section{Personal Attribute Tasks}\label{sec:pa_task}

Based on the usefulness of personal attributes for dialogue personalization, we propose the task of obtaining personal attributes from natural language sentences. We first explain how we formulate two complementary tasks from DialogNLI data and then formally define our tasks. Finally, we analyze the task datasets to gather insights into the linguistic phenomena that our tasks involve.

\subsection{Source of Personal Attributes} 

DialogNLI \citep{welleck-etal-2019-dialogue} contains samples of PersonaChat utterances \citep{zhang-etal-2018-personalizing} in English, each paired with a manually annotated personal attribute triple. Each triple consists of a head entity, a relation, and a tail entity. These triples were initially annotated to identify entailing, contradicting and neutral statements within the PersonaChat corpus. For instance, a statement labelled with (I, [favorite\_color], blue) will contradict with another statement labelled with (I, [favorite\_color], green). The three largest groups of relations are: a. \emph{has\_X} (where X = \emph{hobby, vehicle, pet}) b. \emph{favourite\_Y} (where Y = \emph{activity, color, music}) c. \emph{like\_Z} (where Z = \emph{read, drink, movie}).  

\subsection{Extraction and Inference Tasks}\label{sec:subtask_definition}

By re-purposing the DialogNLI dataset, our tasks seek to extract these personal attribute triples from their paired utterances. We first used a script that obtains pairs of personal triples and utterances. Next, we combined relations with similar meanings such as \textit{like\_food} and \textit{favourite\_food} and removed under-specified relations such as \textit{favourite, have} and \textit{others}. Finally, we removed invalid samples with triples containing \textit{None} or \textit{<blank>} and removed prefix numbers of tail entities (\emph{e.g.} 11 dogs), since the quantity is not important for our investigation.

We formulate two tasks by partitioning the DialogNLI dataset into two non-overlapping subsets. Here, each \textbf{sample} refers to a sentence paired with an annotated triple. Train/dev/test splits follow DialogNLI, with descriptive statistics shown in Table \ref{tab:descriptive}. The dataset for the \textsc{Extraction} task contains samples in which both the head and tail entities are spans inside the paired sentence. An example is (I, [has\_profession], receptionist) from the sentence ``I work as a receptionist in my day job''. We formulate the \textsc{Extraction} task in a similar way to existing Relation Extraction tasks such as ACE05 \citep{wadden-etal-2019-entity} and NYT24 \citep{Nayak_Ng_2020}.
This allows us to apply modeling lessons learned from Relation Extraction.

The complementary set is the dataset for the \textsc{Inference} task, for which the head entity and/or the tail entity cannot be found as spans within the paired sentence. This is important in real-world conversations because people do not always express their personal attributes explicitly and instead use paraphrasing and commonsense reasoning to do so. 
An example of a paraphrased triple is (I, [physical\_attribute], tall) from the sentence ``I am in the 99th height percentile'', while one based on commonsense reasoning is  (I, [want\_job], umpire) from the sentence ``my ultimate goal would be calling a ball game''. 

\begin{table}[t]
\centering
\begin{adjustbox}{max width=\columnwidth}
\begin{tabular}{lcc}
\toprule
& \textsc{Extraction} & \textsc{Inference} \\
\midrule
\multicolumn{3}{l}{\textbf{Samples}}  \\
train & 22911 & 25328\\ 
dev. & 2676 & 2658 \\
test & 2746 & 2452 \\

\midrule
\multicolumn{3}{l}{\textbf{Unique elements}} \\
head entities & 88 & 109\\
relations & 39 &  39 \\
tail entities & 2381 & 2522 \\
\midrule
\multicolumn{3}{l}{\textbf{Avg. words}} \\
head entities & 1.03 & 1.08\\
relations &1.00 & 1.00 \\
tail entities & 1.20 & 1.28 \\
sentences & 12.9 & 12.2 \\

\bottomrule
\end{tabular}
\end{adjustbox}
\caption{Statistics of the dataset for the two tasks. 
}
\label{tab:descriptive}
\end{table}

The \textsc{Inference} task is posed as a challenging version of the \textsc{Extraction} task that tests models' ability to identify pertinent information in sentences and then make commonsense inferences/paraphrases based on such information. An existing task has sought to predict personal attributes that are not always explicitly found within sentences \citep{WuGetting2019}. However, it did not distinguish between personal attributes that can be explicitly found within sentences (\emph{i.e.} \textsc{Extraction}) from those that cannot (\emph{i.e.} \textsc{Inference}) . We believe that, given that the inherent difficulty of identifying the two types of personal attributes are greatly different, it is helpful to pose them as two separate tasks. In this way, the research community can first aim for an adequate performance on the simpler task before applying lessons to make progress at the more challenging task. This is also the first time that personal attributes that are not explicitly contained in sentences are shown to be derivable from words in the sentence using commonsense/lexical inferences.

\subsection{Formal Task Definition}

Given a sentence $S$, we want to obtain a personal-attribute triple in the form of \texttt{(\textbf{head entity, relation, tail entity})}. The relation must belong to a set of 39 predefined relations. In the \textsc{Extraction} subset, the head entity and tail entity are spans within $S$. Conversely, in the \textsc{Inference} subset, the head entity and/or the tail entity cannot be found as spans within $S$.

\subsection{Dataset Analysis}

We analyze the datasets to obtain insights into how the tasks can be approached. Because the majority of head entities (93.3\%) are simply the word ``I'', our analysis will focus on tail entities. 

\paragraph{Dataset for the \textsc{Extraction} task}

\begin{figure}[t]
        \centering
        \includegraphics[width=8cm]{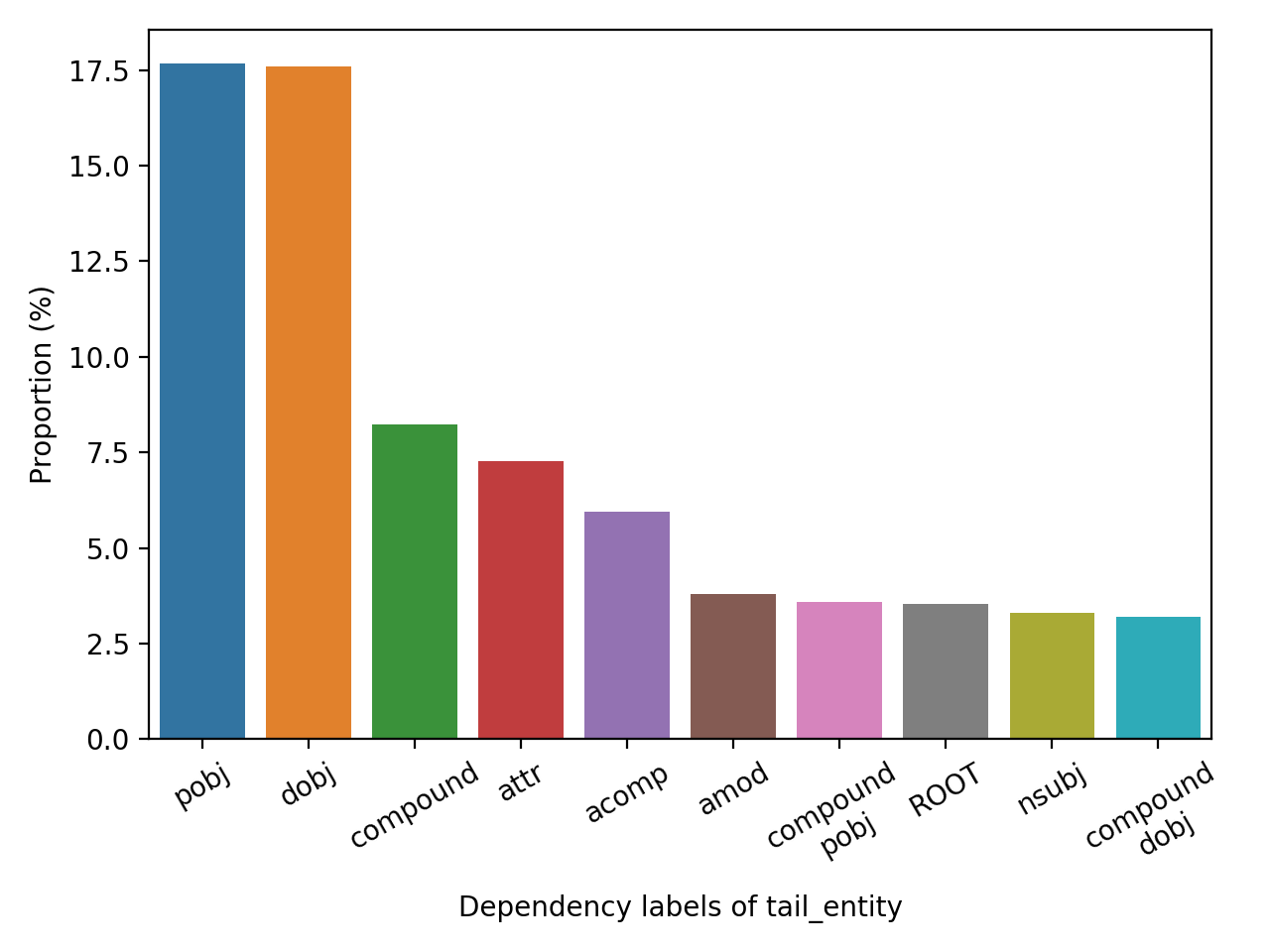}
        \caption{Bar plot for 10 most common dependency role labels of tail entities within sentences}
        \label{fig:barplot_dependency_labels}
\end{figure}

We use dependency parses of sentences to understand the relationship between words within tail entities and the sentence ROOT. Dependency parsing was chosen because it is a well-studied syntactic task \citep{nivre-etal-2016-universal} and previously used for the relation extraction task \citep{zhang2017tacred}. Dependency parses and labels associated with each dependent word were identified using a pre-trained transformer model from spaCy.\footnote{https://spacy.io/} The parser was trained on data annotated with the ClearNLP dependency schema that is similar to Universal Dependencies \citep{nivre-etal-2016-universal}.\footnote{https://github.com/clir/clearnlp-guidelines/blob/master/md/specifications/dependency\_labels.md} 

As shown in Figure \ref{fig:barplot_dependency_labels}, objects of prepositions (pobj) and direct objects (dobj) each comprise 17.5\% of tail entities, followed by compound words (compound), attributes (attr) and adjectival complements (acomp), plus 138 other long-tail labels. The range of grammatical roles as well as the fact that one third of tail entities do not involve nouns (see Figure in Section \ref{sec:task_analysis_details}) also suggest that the tail entities in our dataset go beyond proper nouns, which are what many Relation Extraction datasets (\emph{e.g.}, ACE05 and NYT24) are mainly concerned with. Such diversity in grammatical roles played by tail entities means that approaches based on rule-based extraction, parsing or named entity recognition alone are unlikely to be successful in the \textsc{Extraction} task.




\paragraph{Dataset for the \textsc{Inference} task}

A qualitative inspection of the dataset showed that inferences can be made on the basis of semantically-related words and commonsense inferences, as shown in examples discussed in Section \ref{sec:subtask_definition}.
To better understand how tail entities can be inferred from the sentence in the \textsc{Inference} subset, we analyze the relationship between words in the tail entity and words in the sentence. 79.2\% of tail entities cannot be directly identified in the sentence. We performed a few transformations to identify potential links between the tail entity and the sentence. \textit{ConceptNet\_connect} refers to words with highest-weighted edges on ConceptNet to sentence words while \textit{ConceptNet\_related} refers to words that have closest embedding distances to sentence words. Details of their preparation are in Appendix \ref{sec:transformation_tail_entity}. As in Table \ref{tab:transformation_tail_entity}, our analysis shows that a model that can perform well on the \textsc{Inference} task requiring both WordNet semantic knowledge \citep{wordnet} as well as ConceptNet commonsense knowledge   \citep{10.5555/3298023.3298212}. 


\begin{table}[t]
\centering
\begin{adjustbox}{max width=\columnwidth}
\begin{tabular}{lll}
\toprule
\textbf{Transformation} & \textbf{Example} & \textbf{\%} \\
&\multicolumn{2}{l}{(sentence\textrightarrow tail entity)}\\
\midrule
ConceptNet\_related & mother \textrightarrow female & \textbf{71.3} \\
ConceptNet\_connect & wife \textrightarrow married & 56.8 \\
WordNet\_synonym & outside \textrightarrow outdoors & 39.5 \\
WordNet\_hypernym & drum \textrightarrow instrument & 5.04\\
WordNet\_hyponym & felines \textrightarrow cats & 4.17\\
Same\_stem & swimming \textrightarrow swim & 43.3\\

\bottomrule
\end{tabular}
\end{adjustbox}
\caption{Proportion (\%) of tail entities that can be related to sentence words after applying each transformation.}
\label{tab:transformation_tail_entity}
\end{table}



\section{GenRe}\label{sec:genre}

This section proposes GenRe, a model that uses a unified architecture for both the \textsc{Extraction} and the \textsc{Inference} tasks. We use a simple and extensible generator-reranker framework to address the needs of the two tasks. On one hand, a generative model is necessary because head and/or tail entities cannot be directly extracted from the sentence for the \textsc{Inference} dataset. On the other hand, preliminary experiments using a Generator in isolation showed that a large proportion of correct triples are among the top-k - but not top-1 - outputs. A Reranker can be used to select the most likely triple among the top-k candidate triples, leading to a large improvement in performance (see Table \ref{tab:ablation}).

\subsection{Generator}

We use an autoregressive language model (GPT-2 small) as our Generator because its extensive pre-training is useful in generating syntactically and semantically coherent entities. The small model was chosen to keep model size similar to baselines. We finetune this model to predict a personal attribute triple occurring in a given input sentence. 
Specifically, we treat the flattened triples as targets to be predicted using the original sentence as context. The triple is formatted with control tokens to distinguish the head entity, relation, and tail entity as follows:

$\mathbf{y}$ = [HEAD], $t_{1:m}^{head}$, [RELN], $t^{reln}$,  [TAIL], $t_{1:k}^{tail}$ 

\noindent where \{[HEAD],[RELN], [TAIL]\} are control tokens, $t_{1:m}^{head}$ is the head entity (a sequence of length $m$),  $t^{reln}$ is a relation, and $t_{1:k}^{tail}$ is the tail entity. 

During evaluation, we are given a sentence as context and seek to generate a personal attribute triple in the flattened format as above. To reduce the search space, we adopt a constrained generation approach. Specifically, after the [RELN] token, only one of 39 predefined relations can be generated, and so the output probability of all other tokens is set to 0. 
After the [TAIL] token, all output tokens not appearing in the input sentence will have zeroed probabilities in the \textsc{Extraction} task. Conversely for the \textsc{Inference} task, the only allowed output tokens after the [TAIL] token are those which have appeared following the predicted relation in the training data. For example, tail entities that can be generated with a [physical\_attribute] relation include ``short'', ``skinny'' or ``wears glasses'', as these examples occur in the training data. We imposed this restriction to prevent the model from hallucinating attributes that are not associated to the predicted relation (such as ``dog'' with [physical\_attribute]). Despite limiting the model's ability to generate novel but compatible tail entities (and thereby upper-bounding maximum possible recall to 75.7\%), this approach helped to improve model performance overall. Implementation details are in Appendix  \ref{sec:base_model_details}.

\subsection{Reranker}

We use BERT-base as the Reranker because its bi-directionality allows tail tokens to influence the choice of relation tokens. Furthermore, BERT has demonstrated the best commonsense understanding among pre-trained language models \citep{ petroni-etal-2019-language, Zhou_Zhang_Cui_Huang_2020}.
These benefits have led to many relation extraction models using BERT as part of the pipeline \citep{wadden-etal-2019-entity,yu-etal-2020-dialogue, ye2021contrastive}. 

For each $S$, we obtain the $L$ most likely sequences using the Generator, including the context sentence. Each sequence is labelled as correct or incorrect based on whether the predicted triple (head entity, relation, tail entity) matches exactly the ground-truth triple. Incorrect sequences serve as challenging negative samples for the Reranker because they are extremely similar to the correct sequence since they were generated together. We fine-tune a BERT model with a binary cross-entropy loss function to classify whether sequences are correct. During inference, we select the sequence with the highest likelihood of being correct as our predicted sequence. We set $L$ to 10 in all experiments. Implementation details are in Appendix \ref{sec:discriminator_model_details}. 

\section{Experiments}\label{sec:experiments}

We first explain the metrics used in the experiments. Next, we introduce the baseline models. Finally, we examine how GenRe compares to baseline models to understand its advantages.

\subsection{Metrics}


Micro-averaged Precision/Recall/F1 were calculated following \citet{Nayak_Ng_2020}, in which a sample is considered correct only when all three elements (head\_entity, relation and tail entity) are resolved correctly. We chose these metrics because we are interested in the proportion of all predicted personal attributes that have been correctly identified (precision) and of all ground truth personal attributes (recall). F1 is considered as an aggregate metric for precision and recall. 


\subsection{Baseline Models}


\paragraph{Generative models} can be used for both the \textsc{Extraction} and the \textsc{Inference} tasks.

\textbf{WDec} is an encoder-decoder model that achieved state-of-the-art performance in the NYT24 and NYT29 tasks \citep{Nayak_Ng_2020}. The encoder is a Bi-LSTM, while the decoder is 
an LSTM with attention over encoder states. An optional copy mechanism can be used: when used, the decoder will only generate tokens found in the original sentence. The copy mechanism was used on the \textsc{Extraction} dataset but not on the \textsc{Inference} dataset (given their better empirical performance). 

\textbf{GPT2} is an autoregressive language model that we build GenRe on. We use the same configuration as in GenRe.
\paragraph{Extractive models} can be used only for the \textsc{Extraction} task, because they select for head and tail entities from the original sentence.

\textbf{DyGIE++} is a RoBERTa-based model that achieved state-of-the-art performance in multiple relation extraction tasks including ACE05 \citep{wadden-etal-2019-entity}. It first extracts spans within the original sentence as head and tail entities. Then, it pairs up these entities with a relation and passes them through a graph neural network, with the head and tail entities as the nodes, and relations as the edges. This allows information flow between related entities before passing the triple through a classifier. 



\textbf{PNDec} is an Encoder-Decoder model that achieved close to SOTA performance in NYT24 and NYT29 \citep{Nayak_Ng_2020}. It uses the same encoder as WDec but uses a pointer network to identify head and tail entities from the original sentence, which it pairs with possible relation tokens to form a triple that is subsequently classified.

All baseline models were trained on our datasets using their suggested hyper-parameters.

\subsection{Model Results}

\begin{table}[!h]
\centering
\begin{adjustbox}{max width=\columnwidth}
\begin{tabular}{llll|lll} 
\toprule
& \multicolumn{3}{c|}{\textsc{Extraction}} & \multicolumn{3}{c}{\textsc{Inference}} \\ 
& P & R & F1   & P & R & F1 \\ 
\midrule
GenRe & \textbf{68.0} & \textbf{52.4} & \textbf{59.2} & \textbf{46.5} & \textbf{35.4} & \textbf{39.2}\\ \hdashline

\multicolumn{3}{l}{\textit{Generative}}  & \\
    WDec & 57.0 & 49.0 & 52.7 & 33.6 & 34.7 & 34.1\\
    GPT2 & 50.9 & 31.1 & 38.6 & 31.3 & 17.3 & 22.3\\ 
    
\hdashline
\multicolumn{3}{l}{\textit{Extractive}} & \\
    DyGIE++ & 60.8 & 50.9 & 55.3 & && \\
    PNDec & 63.1 & 49.5 & 55.5 &&& \\

\bottomrule
\end{tabular}
\end{adjustbox}
\caption{Performance on the test set. GenRe has significantly higher mean F1 than all baseline models with 5 runs based on a two-tailed t-test (p $<$ 0.05).}
\label{tab:overall_performance}
\end{table}


The top-performing baseline models on the \textsc{Extraction} dataset are the extractive models, which select spans within the sentence and classify whether an entire triple is likely to be correct. Because there are only a small number of spans within the sentence, this approach can effectively limit its search space. On the other hand, extractive models cannot solve the \textsc{Inference} task, because the underlying assumption that head and tail entities must be found within the sentence does not hold. Conversely, generative models perform more poorly on the Extraction task but are capable on the \textsc{Inference} task. This is because generation happens in a left-to-right manner, meaning that some elements of the triple have to be generated without knowing what the other elements are. Our approach of linking a Generative model with a BERT-base Reranker (akin to models used by Extractive approaches) combines the best of both worlds. Not only does it perform well on the Extraction task ($\geq$ 3.7 F1 points over baselines), it also excels on the Inference task ($\geq$ 5.1 F1 points over baselines).

\section{Analysis}\label{sec:overall_analysis}

We first conduct an ablation study to better understand the contribution of constrained generation and the Reranker, by measuring the performance of our model when each component is removed. Then, we seek to understand how errors are made on predicted personal attribute relations to identify future areas of improvement. 

\subsection{Ablation Study}

Table \ref{tab:ablation} shows that both the Reranker and constrained generation contribute to the performance of GenRe. In particular, the constrained generation plays a larger role on the \textsc{Extraction} dataset while the Reranker plays a greater role on the \textsc{Inference} dataset. 

\textbf{Constrained generation} has a large impact on the \textsc{Extraction} dataset (+13.0\% F1), likely because it very much restricts the generation search space to spans from the context sentence. 
On the \textsc{Inference} dataset, the original search space cannot be effectively limited to tokens in the context sentence. Therefore, applying the heuristic that only tail entities associated with a particular relation (in the training set) can be decoded is useful, even though it upper bounds maximum recall to 75.7\%, which is much higher than the achieved 35.4\%. Compared to the \textsc{Extraction} dataset, the improvement on the \textsc{Inference} dataset is smaller (+7.8\% F1), since the range of tail entities that can be decoded after imposing the constraint is greater.

\begin{table}[!t]
\centering
\begin{adjustbox}{max width=\columnwidth}
\begin{tabular}{llll|lll} 
\toprule
& \multicolumn{3}{c|}{\textsc{Extraction}} & \multicolumn{3}{c}{\textsc{Inference}} \\ 
& P & R & F1   & P & R & F1 \\ 
\midrule
 
 GenRe & 68.0 & 52.4 & 59.2 & 46.5 & 35.4 & 39.2\\ 
- Constr. Gen & 53.5 & 40.7	& 46.2 & 37.2 & 27.1 & 31.4\\
- Reranker  & 67.6 & 41.0 & 51.0 &  31.0 & 22.3 & 25.9\\

\bottomrule
\end{tabular}
\end{adjustbox}
\caption{Ablation study for Reranker and constrained generation.}
\label{tab:ablation}
\end{table}

\begin{table*}[!h]
\centering
\begin{adjustbox}{max width=\textwidth}
\begin{tabular}{l|llll|l|l|l}
\toprule
&&&&&  \multicolumn{3}{c}{Top 3 Most Frequent (n)} \\ 
Dataset & True Relation (n) & P & R & F1 &Predicted Relations  & True Tail Entities & Predicted Tail Entities\\
\midrule
 \textsc{Extraction}  & {[has\_profession]} (274) & 83.8 & 62.0 & 71.3 & {[has\_profession]} (189) & teacher (29) & nurse (27) \\
&&&& & {[employed\_by\_general]} (30) & nurse (28) & real estate (25) \\
&&&& & {[want\_job]} (17) & real estate agent (25) & teacher (19) \\
\cline{1-8}
 & {[have\_pet]} (149) & 97.3 & 55.0 & 70.3 & {[have\_pet]} (88) & dog (55) & cat (32) \\
&&&& & {[have\_family]} (18) & cat (45) & pets (23) \\
&&&& & {[like\_animal]} (12) & pets (22) & dog (18) \\
\midrule
\textsc{Inference} & {[like\_food]} (77) & 46.7 & 41.6 & 44.0 & {[like\_food]} (62) & pizza (18) & pizza (19) \\
&&&& & {[like\_activity]} (5) & onion (9) & italian cuisine (10) \\
&&&& & {[like\_animal]} (4) & italian (7) & onion (8) \\

\cline{2-8}
& {[like\_music]} (71) & 40.8 & 23.9 & 30.2 & {[like\_music]} (40) & jazz (10) & the story so far (12) \\
&&&& & {[favorite\_music\_artist]} (9) & country (9) & country (8) \\
&&&& & {[like\_activity]} (7) & rap (6) & jazz (7) \\

\bottomrule
\end{tabular}
\end{adjustbox}
\caption{Some relations in \textsc{Extraction} and \textsc{Inference} datasets} 
\label{tab:f1_by_relation}
\end{table*}

\textbf{The Reranker} is needed because, many times, the correct triple can be generated by the Generator but might not be the triple that is predicted to have the highest likelihood. 
The maximum possible recall on the \textsc{Extraction} and \textsc{Inference} tasks increases from 41.0\% to 59.9\% and 22.3\% to 41.0\% respectively when considering top-10 instead of only top-1 generated candidate. While the achieved recall (52.4\% and 35.4\% respectively) are still a distance from the maximum possible recall, the achieved recall is much higher than using the Generator alone.

\subsection{Misclassification of Relations}

Major sources of error on the \textsc{Extraction} dataset came from relation tokens that have close semantic meanings. They either were related to one another (\emph{e.g.}, [has\_profession] vs [want\_job]) or could be correlated with one another (\emph{e.g.}, [like\_animal] vs [have\_pet] or [like\_music] vs [favorite\_music\_artist]) , as illustrated in Table \ref{tab:f1_by_relation}. Such errors likely arose due to the way that the DialogNLI dataset \citep{welleck-etal-2019-dialogue} was annotated. Specifically, annotators were asked to label a single possible triple given a sentence instead of all applicable triples. Because of this, our evaluation metrics are likely to over-penalize models when they generate reasonable triples that did not match the ground truth. Future work can avoid this problem by labelling all possible triples and framing the task as multilabel learning.

\section{Applications of Personal Attributes}\label{sec:benefits_to_personachat}

Personal attributes can make social chit-chat agents more consistent and engaging as well as enable task-oriented agents to make personalized recommendations. In this section, we use personal attributes to improve chit-chat agent consistency and provide information for personalizing task-oriented dialogue agents.





\subsection{Consistency in Chit-chat agents}
PersonaChat \citep{zhang-etal-2018-personalizing} was created to improve the personality consistency of open-domain chit-chat dialogue agents. PersonaChat was constructed by giving pairs of crowdworkers a set of English personal attribute related sentences and asking them to chat in a way that is congruent with those sentences. Models were then trained to generate dialogue responses that are in line with those expressed by crowdworkers using the provided persona information as context. 

\textbf{Methods} We fine-tune the generative version of Blender 90M  (a transformer-based model trained on multiple related tasks) on PersonaChat, which is currently the state-of-the-art generative model on this task \citep{roller2020recipes} and uses personal attribute sentences to ground dialogue response generation. Building on Blender, we prepend a corresponding DialogNLI personal attribute before each utterance (\emph{i.e.} \textbf{+Per. Attr.}), in order to better direct the model in generating a suitable response that is consistent with the set persona. This modification is relatively minimal to demonstrate the informativeness of personal attribute KG triples, while keeping the model architecture and hyperparameter fine-tuning the same as in the original work (details in Appendix \ref{sec:blender_details}). 

\textbf{Metrics} We follow \citet{roller2020recipes} and \citet{dinan2019second}. Metrics for \textbf{+Per. Attr.} setting consider both personal attributes and utterances. 
\textbf{Hits@1} uses the hidden states of the generated output to select the most likely utterance amongst 20 candidates (the correct utterance and 19 randomly chosen utterances from the corpus). \textbf{Perplexity} reflects the quality of the trained language model. \textbf{F1} demonstrates the extent of the overlap between the generated sequence and the ground truth sequence.  


\begin{table}[!h]
\centering
\begin{adjustbox}{max width=\columnwidth}
\begin{tabular}{llll}
\toprule
& Hits@1 $\uparrow$ & Perplexity $\downarrow$ & F1 $\uparrow$ \\
\midrule
Blender & 32.3 & 11.3  & 20.4 \\ 
+ Per. Attr. & 35.2*  & 10.4*  & 20.6* \\

\bottomrule
\end{tabular}
\end{adjustbox}
\caption{Effects of using personal attributes to augment Blender on Personachat.  Higher is better for Hits@1 and F1; lower is better for perplexity. *Significantly different from Blender with 5 runs based on a two-tailed t-test (p$<$0.05). }
\label{tab:application_results}
\end{table}

\begin{table}[!h]
\centering
\begin{adjustbox}{max width=\columnwidth}
\begin{tabular}{lll}
\toprule
\textbf{Fact 1} & I love cats and have two cats\\ 
\textbf{Fact 2} & I've a hat collection of over 1000 hats. \\
\textbf{Blender} & My cats names are all the hats i have \\
\textbf{+ Per. Attr.} & My cats are called kitties \\
\midrule
\textbf{Fact 1} & I am a doctor. \\
\textbf{Fact 2} & My daughter is a child prodigy.\\
\textbf{Blender} & My daughter is prodigy so she gets a lot of accidents.\\
\textbf{+ Per. Attr.} & I've seen a lot of accidents.\\

\bottomrule
\end{tabular}
\end{adjustbox}
\caption{Examples of incorrect utterances generated by Blender by mixing up two facts, which are avoided by our Blender + Per. Attr. model}
\label{tab:examples_of_blender_mistakes}
\end{table}

\textbf{Results} As shown in Table \ref{tab:application_results}, including personal attributes can improve performance on the PersonaChat task. An inspection of the generated utterances suggests that including personal attributes into Blender can more effectively inform the model which persona statement to focus on during generation. This can prevent Blender from including information in irrelevant persona statements (\emph{e.g.} by mixing up facts from two unrelated persona statements), as in Table \ref{tab:examples_of_blender_mistakes}. 

\subsection{Personalization in Task-oriented dialogue}

While personalization has been incorporated into single-task settings \citep{joshi2017personalization, mo2017personalizing, Luo_Huang_Zeng_Nie_Sun_2019, lu-etal-2019-goal, pei2021cooperative}, there has been no attempt for personalization in multi-task settings. This is against the background in which multi-task dialogue is rapidly becoming the standard in task-oriented dialogue evaluation \citep{ taskmaster, rastogi2019towards, zang2020multiwoz, shalyminov2020fast}. To overcome this gap, we show how our dataset can lay a foundational building block for personalization in multi-task dialogue.



\textbf{Methods} We used several popular datasets on multi-task task-oriented dialogue \citep{zang2020multiwoz, shalyminov2020fast, taskmaster, rastogi2019towards}. From each dataset, we manually observed its tasks and categorized them into several overarching domains, as shown in Table \ref{tab:domains_dataset}. We then created a mapping between the various domains and datasets available for personalizing task-oriented dialogue (including ours). Domains that are not supported by any dataset are omitted.

\textbf{Results}
Compared to existing datasets in Table \ref{tab:domains_dataset}, our dataset is capable of personalizing recommendations in a much larger number of domains. These domains include restaurants and shopping, which have been explored by existing datasets, as well as movies, music, sports and recreation, which have thus far been overlooked. For domains that have been previously explored, such as restaurants, our dataset contains a more diverse set of possible personal attribute values (\emph{e.g.} the foods people like), which can support it to personalize recommendations in more realistic manners. 

\begin{table}[!t]
\centering
\begin{adjustbox}{max width=\columnwidth}
\begin{tabular}{lll}
\toprule
Dataset & Domains &  \#Unique \\
&& features\\
\midrule
Ours & Restaurants, Movies, & 5583 \\
&  Music, Sports,\\
&  Recreation, Shopping \\
Ours & Restaurants \textit{only} & 206 \\
\citet{joshi2017personalization} & Restaurants & 30\\
\citet{mo2017personalizing} & Restaurants & 10 \\
\citet{lu-etal-2019-goal} & Shopping & 7\\
\bottomrule
\end{tabular}
\end{adjustbox}
\caption{Domains covered by various datasets for personalizing task-oriented dialogue. \#Uniques features refers to the number of unique attribute-values (\emph{e.g.} the specific food people like) that can be used for personalization.}
\label{tab:domains_dataset}
\end{table}

\section{Related Work}\label{sec:related_work}

\textbf{Personal Attribute Extraction:} Most work on extracting personal attributes from natural language \citep{pappu-rudnicky-2014-knowledge, mazare-etal-2018-training, WuGetting2019, tigunova2019listening,tigunova-etal-2020-charm} employed distant supervision approaches using heuristics and hand-crafted templates, which have poor recall. In contrast, we use a strong supervision approach in which triples were manually annotated. \citet{7078578} and \citet{yu-etal-2020-dialogue} attempted to extract personal information from dialogue using a strongly supervised paradigm. However, they focused on demographic attributes as well as interpersonal relationships, which contrast with our focus on what people own and like. \citet{7078578} used SVMs to classify relations and CRFs to perform slot filling of entities while  \citet{yu-etal-2020-dialogue} used BERT to identify relations between given entities.

\hspace{-0.2in} \textbf{Generating KG triple using Language Models:} Autoregressive language models have been applied to a wide range of tasks involving the generation of data with similar structures as personal attribute KG triples, including dialogue state tracking \citep{hosseiniasl2020simple} and commonsense KG completion  \citep{Bosselut2019COMETCT}. The most similar application is \citet{alt2019improving}, which used the original GPT model \citep{Radford2018ImprovingLU} for relation classification. Their task formulation involves identifying a specific relation (out of around 30 possible options) for two given entities. On the other hand, our tasks seek to identify not only the relation, but also the head and tail entities, which have potentially open vocabulary requirements, which makes them much harder. 










\section{Conclusion}

In conclusion, we propose the novel tasks of extracting and inferring personal attributes from dialogue and carefully analyze the linguistic demands of these tasks. To meet the challenges of our tasks, we present GenRe, a model which combines constrained attribute generation and re-ranking on top of pre-trained language models. GenRe achieves the best performance vs.\ established Relation Extraction baselines on the Extraction task ($\geq 3.7$ F1 points) as well as the more challenging \textsc{Inference} task that involves lexical and commonsense inferences ($\geq 5.1$ F1 points). Together, our work contributes an important step towards realizing the potential of personal attributes in personalization of social chit-chat and task-oriented dialogue agents. 

\section*{Acknowledgments}
We thank Noah Smith, Waleed Ammar, Massimilliano Ciaramita, Dongqi Su, Weizhe Lin and many others at UW-NLP and beyond for insight discussions.

\section*{Ethics and Broader Impact}

\textbf{Privacy in real world applications} We inspected a selection of the Dialog NLI dataset to ensure it contains no real names, personally-identifying information or offensive content. Because our task involves extracting and inferring personal attributes, real-world users should be given the option to disallow particular types of relations from being collected and/or used for downstream applications. Users should also be given the freedom to delete their collected personal attributes. A further step might be to restrict the extraction and storage of personal attributes to only local devices using differential privacy and federated learning techniques.

\bibliography{references}

\clearpage

\appendix
\renewcommand\thesection{\Alph{section}}
\section{Appendix}

\subsection{Blender Fine-tuning Details}\label{sec:blender_details}

Finetuning hyperparameters are taken from https://parl.ai/projects/recipes/, with the exception of validation metric changed to Hits@1. Each fine-tuning epoch takes 1.5 hours on a Nvidia V100 GPU. We only prepend personal attributes before system utterances but not user utterances. Metrics are for the validation set because test set was not available. All experiments were conducted using ParlAI \citep{miller2017parlai}.

\subsection{Task Analysis Details}\label{sec:task_analysis_details}

\begin{figure}[h]
        \centering
        \includegraphics[width=8cm]{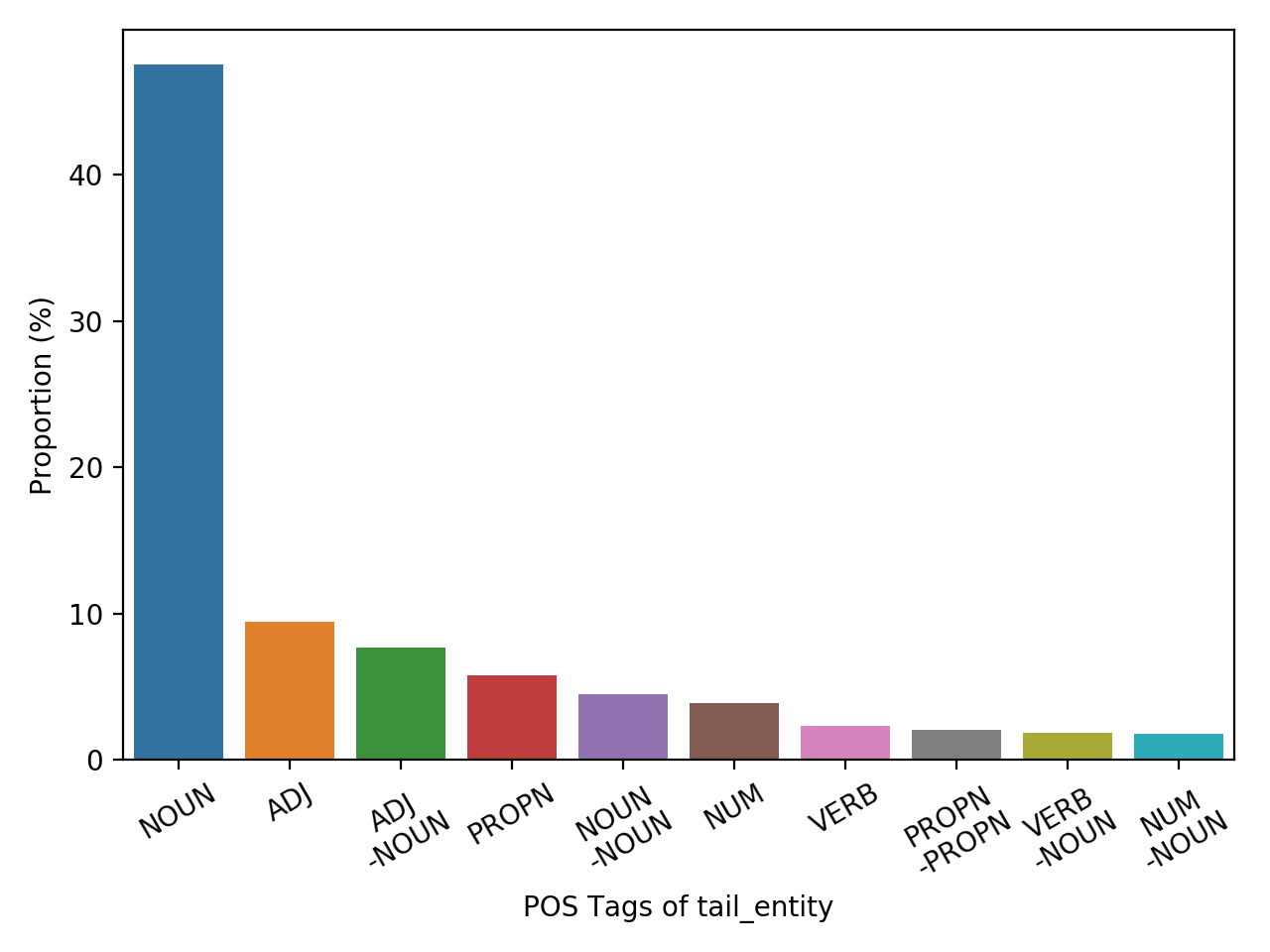}
        \caption{Bar plot for 10 most common POS tags of tail entities.}
        \label{fig:barplot_pos_tags}
\end{figure}

\subsection{Details of Transformations to Link Tail Entity to Sentence}\label{sec:transformation_tail_entity}

\textbf{ConceptNet\_related}: All words in the tail entity can be found in the 100 most related words to each sentence word based on embedding distance on ConceptNet

\textbf{ConceptNet\_connect}: All words in the tail entity can be found in the 100 words that have the highest-weighted edge with each sentence word on ConceptNet.

\textbf{WordNet\_synonym}: All words in the tail entity can be found in the synonyms of every synset of each sentence word on WordNet.

\textbf{WordNet\_hypernym}:  All words in the tail entity can be found in the hypernyms of every synset of each sentence word on WordNet 

\textbf{WordNet\_hyponym}:  All words in the tail entity can be found in the hyponyms of every synset of each sentence word on WordNet 

\textbf{Same\_stem}: All words in the sentence and tail entity are stemmed using a Porter Stemmer \citep{porter1980algorithm} before searching for the tail entity in the sentence

\subsection{Generator Details}\label{sec:base_model_details}

GPT-2-small was used. Additional special tokens including the control tokens ([HEAD], [RELN], [TAIL]) as well as relation tokens were added into the tokenizer. Beam search decoding (beam size = 10) was used at inference time. GPT2-small was accessed from HuggingFace Transformers library with 125M parameters, context window 1024, 768-hidden, 768-hidden, 12-heads, dropout = 0.1. 
AdamW optimizer was used with $\alpha=7.5*10^{-4}$ for the \textsc{Extraction} dataset and  $\alpha=2.5*10^{-3}$ for the \textsc{Inference} dataset, following a uniform search using F1 as the criterion at intervals of $\{2.5, 5,  7.5, 10\} * 10^{n}; -5 \leq n \leq -3$. Learning rate was linearly decayed (over a max epoch of 8) with 100 warm-up steps. Each training epoch took around 0.5 hour on an Nvidia V100 GPU with a batch size of 16. Validation was done every 0.25 epochs during training. 5 different seeds (40-44) were set for 5 separate runs.

\subsection{Reranker Details}\label{sec:discriminator_model_details}

BERT-base-uncased was used. Additional special tokens including the control tokens ([HEAD], [RELN], [TAIL]) as well as relation tokens were added into the tokenizer.  
BERT-base-uncased was accessed from HuggingFace Transformers library (with 12-layer, 768-hidden, 12-heads, 110M parameters, dropout = 0.1). The choice of the base model was made to have fairness of comparison with baseline models in terms of model size.  AdamW optimizer was used with $\alpha=5*10^{-6}$, following a uniform search using F1 as the criterion at intervals of $\{2.5, 5,  7.5, 10\} * 10^{n}; -6 \leq n \leq -3$. Learning rate was linearly decayed (over a max epoch of 8) with 100 warm-up steps. Each training epoch took around 1 hour on an Nvidia V100 GPU with a batch size of 10.Validation was done every 0.25 epochs during training. 5 different seeds (40-44) were set for 5 separate runs.

\newpage

\end{document}